\pdfoutput=1

\documentclass[11pt]{article}

\usepackage[final]{acl}

\usepackage{times}
\usepackage{latexsym}
\usepackage{algorithm}
\usepackage{algpseudocode}

\usepackage{bbm}
\usepackage[para]{threeparttable}
\usepackage{booktabs}
\usepackage{multirow}
\usepackage{amsmath}
\usepackage{amssymb}
\usepackage{xcolor}
\usepackage{hyperref}
\usepackage{mathtools}
\usepackage{tcolorbox}
\usepackage{listings}
\usepackage{enumitem}
\usepackage{makecell}
\definecolor{c1}{HTML}{0049C0}
\usepackage{colortbl}
\usepackage{tabularx}
\usepackage{pifont}

\definecolor{mycell}{rgb}{0.85, 0.93, 0.97}
\definecolor{mycelltwo}{RGB}{255, 238, 241}
\definecolor{Ground}{RGB}{255,184,55}
\definecolor{Rice}{RGB}{251,248,238}
\definecolor{Dirt}{RGB}{191,169,115}
\definecolor{Pink}{RGB}{226,184,176}
\definecolor{Violet}{RGB}{163,148,170}
\definecolor{mygray}{RGB}{226, 226, 226}
\usepackage{graphicx}

\usepackage[T1]{fontenc}

\usepackage[utf8]{inputenc}

\usepackage{microtype}

\usepackage{inconsolata}

\usepackage{graphicx}

\definecolor{model1}{RGB}{240, 29, 5}
\definecolor{model2}{RGB}{5, 95, 240}

\newcommand{\fname}{\textsc{CHIQ}}

\newcolumntype{g}{>{\columncolor{Ground!10}}c}
\newcolumntype{d}{>{\columncolor{Dirt!10}}c}
\newcolumntype{f}{>{\columncolor{Pink!10}}c}
\newcolumntype{v}{>{\columncolor{Violet!10}}c}
\newcolumntype{P}[1]{>{\centering\arraybackslash}p{#1}}

\title{ICR: Iterative Clarification and Rewriting for Conversational Search}

\author{Zhiyu Cao, Peifeng Li\thanks{ \ \ Corresponding author}, Qiaoming Zhu \\
        School of Computer Science and Technology, Soochow University, Suzhou, China  \\
        \texttt{zycao18@stu.suda.edu.cn}, \texttt{\{pfli, qmzhu\}@suda.edu.cn}
        }

\begin{document}
\maketitle
\begin{abstract}
Most previous work on Conversational Query Rewriting employs an end-to-end rewriting paradigm. However, this approach is hindered by the issue of multiple fuzzy expressions within the query, which complicates the simultaneous identification and rewriting of multiple positions. To address this issue, we propose a novel framework ICR (Iterative Clarification and Rewriting), an iterative rewriting scheme that pivots on clarification questions. Within this framework, the model alternates between generating clarification questions and rewritten queries. The experimental results show that our ICR can continuously improve retrieval performance in the clarification-rewriting iterative process, thereby achieving state-of-the-art performance on two popular datasets.
\end{abstract}

\section{Introduction}
\begin{figure}[t]
\begin{center}
 \includegraphics[width=1\linewidth]{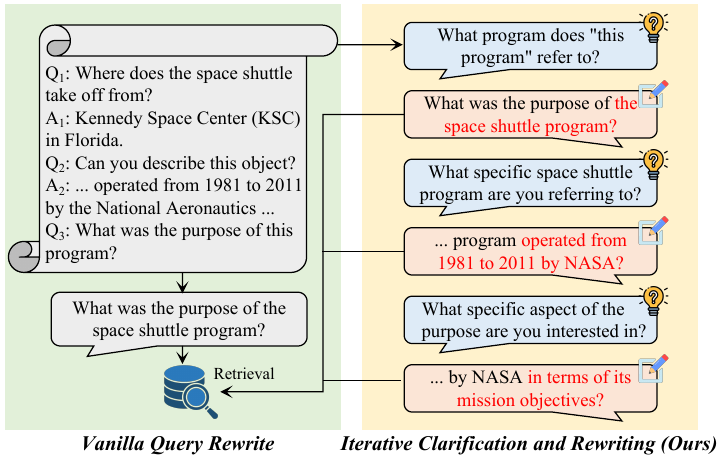}
 \caption{ICR differs from those traditional query rewriting methods in that we use clarification questions as the pivot to guide rewriting.}
 \label{fig:icr_comp}
\end{center}
\vspace{-0.5cm}
\end{figure}

In conversational question answering, users' questions often require the assistance of external knowledge. Conversational search is designed to provide users with external information needs in multi-turn conversation.  However, users often omit some content in their queries for the sake of simplification during the conversation. Conversational Query Rewriting (CQR) is a key step in conversational search, aiming to rewrite vague queries in conversation into de-contextualized queries, thereby promoting conversational search. As shown in Figure~\ref{fig:icr_comp}, given the dialogue history (i.e., $Q_1$ to $A_2$) and the current query $Q_3$, the goal of CQR is to rewrite $Q_3$ into a more complete query (e.g., ``What was the purpose of the space shuttle program operated from 1981 to 2011 by NASA?'') and use this rewritten query to retrieve relevant passages.

Early studies on CQR used manually labeled rewritten queries as ground truth to train the models~\cite{T5QR,CONQRR,EDIRCS,ConvGQR}. However, manually labeled rewritten queries often only conform to human readability and may not necessarily be conducive to conversational search. Therefore, some recent studies~\cite{LLM-Aided,LLM4CS,IterCQR,CHIQ,AdaCQR,RETPO} have explored using LLMs to generate rewritten queries.
Although these studies have achieved better retrieval performance, they still employed an end-to-end paradigm, directly generating rewritten query based on the dialogue history and current query. As with humans, LLMs do not always produce the best output on the first try~\cite{Self-refine}. Therefore, the limitation of the above paradigm lies in the fact that the user's query may have multiple potential ambiguities in expression (e.g., $Q_3$ in Figure~\ref{fig:icr_comp} does not specify either what program it is or when the program occurs), and it is difficult to resolve all the ambiguities by rewriting the query in one step, making it challenging to deal with extremely ambiguous queries. This highlights the need for advanced rewrite mechanisms to handle these complex scenarios.

To address the above issues, we propose a novel framework ICR (Iterative Clarification and Rewriting). ICR first constructs iterative clarification-rewriting data based on the retrieval performance of the query. The model then learns to engage in the process of clarification-rewriting, alternately generating clarification questions and rewritten queries. This process is critical to identify and rectify the ambiguites and omissions in the query. As shown in Figure~\ref{fig:icr_comp}, this strategy advances the traditional CQR methods by explicitly decomposing the query rewriting process into multiple iterations that include clarification and rewriting, each designed to progressively refine the rewritten queries. Since multiple queries are generated during the iteration process, we fused the retrieval results based on the contribution of the rewritten queries in the iterative process. The experimental results on TopiOCQA and QReCC show that our ICR can achieve SOTA performance.

\begin{figure*}[t]
\begin{center}
 \includegraphics[width=1\linewidth]{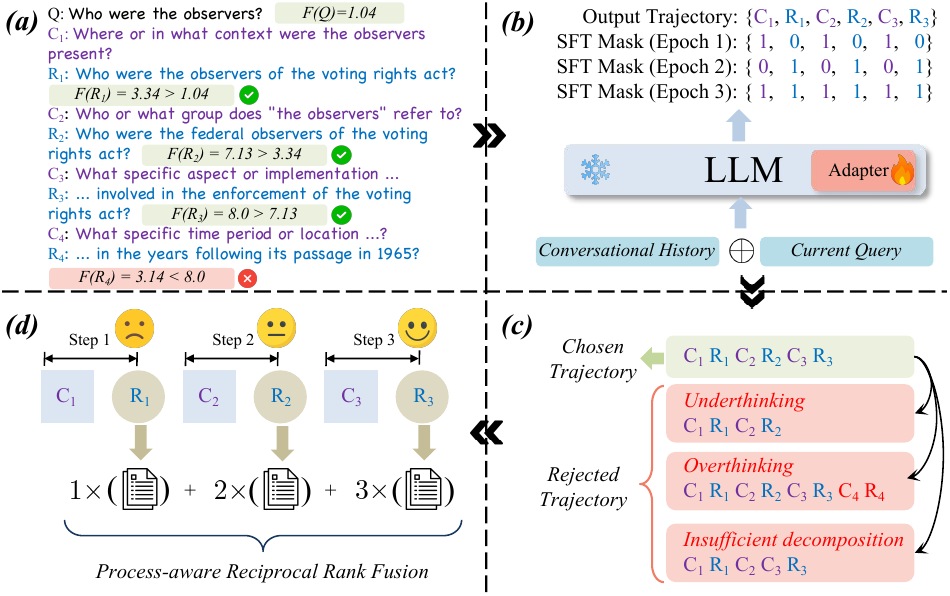}
 \vspace{-24pt}
 \caption{\textbf{Overview of the ICR framework.} ICR consists of four main modules: (a) CRDG: Clarification-Rewriting Data Generation (\autoref{sec:data_con}), (b) PSFT: Progressive Supervised Fine-Tuning from Clarification to Rewriting (\autoref{sec:progressive_sft}), (c)  DPO-CRP: Direct Preference Optimization for Clarification-Rewriting Process (\autoref{sec:process_dpo}), and (d) PRRF: Process-aware Reciprocal Rank Fusion (\autoref{sec:prrf}).}
 \label{fig:method}
\end{center}
\vspace{-12pt}
\end{figure*}

\section{Related Work}
Previous work used manually labeled rewritten queries as supervisory signals. ~\citet{T5QR} introduced pre-trained language models to relax the independence assumptions made when using MLE objective in the CQR task. CONQRR~\cite{CONQRR} used reinforcement learning to adjust and optimize against off-the-shelf retrievers for conversational retrieval. Since previous studies generated tokens one by one from scratch, EDIRCS~\cite{EDIRCS} simultaneously selected the tokens from dialogues and generated a portion of new tokens, and then proposed two conversational search-oriented learning objectives. ConvGQR~\cite{ConvGQR} proposed a knowledge fusion mechanism combining query rewriting and expansion.

Since manually labeled rewritten queries are often not optimal, recent work has focused on using LLMs to generate rewritten queries. ~\citet{LLM-Aided} defined four important properties for rewritten queries and used LLMs as rewrite editors. LLM4CS ~\cite{LLM4CS} explored three prompting methods to generate rewritten queries. By using the signal of information retrieval as a reward, IterCQR 
 ~\cite{IterCQR} did not rely on manually annotated rewritten queries. CHIQ ~\cite{CHIQ} enhanced the conversation history and then generated rewritten queries based on the enhanced conversation history through three approaches. Considering the two perspectives of the term and semantics, AdaCQR ~\cite{AdaCQR} introduced the contrastive loss to optimize the reformulation model. RETPO~\cite{RETPO} optimized a language model to generate rewritten queries preferred by the retriever.

Unlike previous studies directly using manually labeled or LLM-labeled rewritten queries as supervised signals, our ICR decomposes CQR into several iterations of clarification-rewriting, dynamically rewriting the current query and continuously refine it during iterations.

\section{Methodology}

\subsection{Problem Formulation}
Conversational search is to find the top-$K$ most relevant passages $\mathcal{P}_t=\left\{p_j\right\}_{j=1}^K$ from a large collection $\mathcal{C}$ ($p_j \in \mathcal{C}$) based on the current query $q_{t}$ and the dialogue history $\mathcal{H}_{t-1}=\left\{q_i, a_i\right\}_{i=1}^{t-1}$ where $q_i$ and $a_i$ are the $i$-th query and response, respectively. These passages are then used to generate the $t$-th turn response $a_t$. The CQR task is commonly employed as an intermediate step to rewrite the ambiguous query $q_{t}$ into a self-contained rewritten query $r_{t}$, which is then used for passage retrieval. The process can be formally represented as follows:
\begin{equation}\label{eq:Formulation}
r_{t} \leftarrow CQR\left([\mathcal{H}_{t-1}; q_{t}]\right), \mathcal{P}_t \leftarrow \mathcal{R}(r_{t}, \mathcal{C}, K),
\end{equation}
where $CQR(\cdot)$ represents the CQR model and $\mathcal{R}(\cdot)$ can be either a sparse or dense retriever. For brevity, we omit the subscripts later and CQR can be represented as $r \leftarrow CQR\left([\mathcal{H}; q]\right)$.

\subsection{Overview}
As shown in Figure~\ref{fig:method}, our ICR consists of four modules. We first construct iterative clarification-rewriting data offline based on the retrieval performance of rewritten queries. Secondly, we propose a progressive fine-tuning scheme, learning to ask clarification questions and rewrite queries based on those clarification questions. Thirdly, we construct preference data for preference alignment from three dimensions: overthinking, underthinking, and insufficient decomposition. Finally, to fully take into account the contribution of the different rewritten queries during the iteration process, we propose process-aware reciprocal rank fusion.

\subsection{Clarification-Rewriting Data Generation}
\label{sec:data_con}
To construct the clarification-rewriting iterative data, we introduce LLMs to simulate the roles of asking clarification questions and rewriting based on the clarification questions. The specific process of Clarification-Rewriting Data Generation (CRDG) is shown in Algorithm~\ref{alg:Clarification-Rewrite_data}. In each iteration, we first use an LLM to generate a clarification question based on the query from the previous iteration, and then generate a new query based on the conversational context and the current clarification question (the prompts for clarification and rewriting are shown in Appendix~\ref{sec:prompt}) as follows:

\algnewcommand\algorithmicforeach{\textbf{for each}}
\algdef{S}[FOR]{ForEach}[1]{\algorithmicforeach\ #1\ \algorithmicdo}

\algrenewcommand\algorithmicensure{\textbf{Step}}

\begin{algorithm}[t]

\caption{Data Construction}\label{alg:Clarification-Rewrite_data}
\begin{algorithmic}[1]
\small
\Require Conversational query rewriting dataset $D_{CQR} = \{\mathcal{H}^{i}, q^{i}, \mathcal{P}^{i}\}$; Early stopping parameter $\mathcal{E}$; Maximum number of clarification iterations $\mathcal{I}$; Large language model $\mathcal{M}$.

\State $D_{cr} \gets \{\}$ \Comment{\textcolor{gray!50}{Initialize the clarification-rewriting dataset}}
\ForEach {$(\mathcal{H}^{i}, q^{i}, \mathcal{P}^{i}) \in D_{CQR}$ }
    \State  $R_{max} \gets 0$, $num_d \gets 0$, $\hat{q} \gets {q}^{i}, \tau \gets $``''
    \For{$e=1$ to $\mathcal{I}$}
        \State $c \gets \mathcal{M}(\hat{q})$ \Comment{\textcolor{gray!50}{Sampling clarification question}}
        \State $r \gets \mathcal{M}(\mathcal{H}^{i},\hat{q},c)$ \Comment{\textcolor{gray!50}{Generate rewritten query}}
        \State Calculate $F(r)$ according to $\mathcal{P}^i$ and Eq. \ref{eq:rs_calculation}.
        \If {$F(r) > R_{max}$}
            \State $R_{max} \gets F(r)$, $num_d \gets 0$
            \State $\hat{q} \gets r$ \Comment{\textcolor{gray!50}{Update rewritten query}}
        \Else
            \State $num_d \gets num_d+1$
            \State Break if $num_d \geq \mathcal{E}$
        \EndIf
        \State Add ``\texttt{[Clarification]} $c$ \texttt{[Rewrite]} $r$'' to $\tau$
    \EndFor
    \State Add $(H^{i}, q^{i}, \tau)$ to $D_{cr}$
\EndFor
\end{algorithmic}

\end{algorithm}
\vspace{-7mm}

\begin{equation}
    c_{i} \sim \mathcal{M}(r_{i-1}), r_i \sim \mathcal{M}(c_{i}, \mathcal{H}, r_{i-1}),
\end{equation}
where $c_{i}$ and $r_{i}$ represent the clarification question and rewritten query generated in the $i$-th turn, respectively. Note that in the first iteration, we use the original query (i.e., $r_0=q$). In this process, we need to determine when to stop asking clarification questions, i.e., generate the final rewritten query. If the quality of the rewritten query does not improve continuously, it should stop asking clarification questions. Here we use the retrieval performance of rewritten query to measure the quality of rewritten query as follows:
\begin{equation}
\vspace{-2mm}
\label{eq:rs_calculation}
    F(r) =\sum\limits_{m \in MT} m_s(r,d_r)+m_d(r,d_r),
\end{equation}
where $MT = \{\text{MRR},\text{NDCG},\text{R@10},\text{R@100}\}$ which is described in Section~\ref{sec:exp}, $m_s$ represents sparse retrieval, $m_d$ represents dense retrieval, and $d_r$ represents the gold passages corresponding to the query $r$. $F(r)$ denotes the sum of the four metrics in $MT$ for both sparse and dense retrieval. A higher $F(r)$ indicates a better quality rewritten query. In the $i$-th iteration, if $F(r_i) \leq F(r_{i-1})$, the clarification and rewriting of this round fails and resampling is performed.

We set the early stopping parameters $\mathcal{E}$ and maximum iteration rounds $\mathcal{I}$. If the retrieval performance does not improve for $\mathcal{E}$ consecutive iterations or has iterated $\mathcal{I}$ rounds, the iteration will be terminated. Based on the above operation, we can collect an iterative clarification-rewriting dataset $D_{cr} = \{x^{(i)},\tau^{(i)}\}$. $x$ is the concatenation of dialogue history $\mathcal{H}$ and the current query $q$, $\tau=(c_1,r_1,\dots, c_n,r_n)$ is the clarification-rewriting iterative trajectory, where $F(r_1) < F(r_2) < \dots < F(r_n)$. In order to distinguish between clarification question and rewritten query, as well as to allow parsing of the model output during inference, we add \texttt{[Clarification]} and \texttt{[Rewrite]} respectively before the clarification question and rewritten query.

\subsection{Progressive Supervised Fine-Tuning from Clarification to Rewriting}
\label{sec:progressive_sft}
Once the iterative clarification-rewriting data $D_{cr}$ is ready, we optimize the model $\pi_\theta$ to initialize the clarification-rewriting behavior. During the clarification-rewriting iterative process, the model needs to learn how to ask clarification questions as well as how to rewrite based on the clarification questions. Since asking clarification questions and rewriting are two different skills, training them together may confuse the model. How to decouple two skills is crucial for the learning of the model.

Therefore, we propose a Progressive Supervised Fine-Tuning (PSFT) approach, consisting of three epochs. In the first and second epochs, we mask out the loss corresponding to the rewritten queries and clarification questions respectively. In the third epoch, no tokens are masked out, i.e., training the model's ability to ask clarification questions and rewrite simultaneously. For the token $t$ in the input $x$, its corresponding mask is $\delta_{mask}(t)$ as follows:
\[
\small
\delta_{mask}(t)=
\begin{cases}
0, & \text{if } \text{Type}(t) = \texttt{Rewrite}\text{ \& Epoch 1}  \\
0, & \text{if } \text{Type}(t) = \texttt{Clarification}\text{ \& Epoch 2}  \\
1, & \text{otherwise}
\end{cases}
\]
where Type($\cdot$) indicates which type the token belongs to, $\text{Type}(t) = \texttt{Rewrite}$ if it corresponds to a rewritten query, or $\text{Type}(t) = \texttt{Clarification}$ if it corresponds to a clarification question. We optimize our policy $\pi_\theta$ by minimizing the following objective:
\begin{equation}
\small
\mathcal{L}=-
\mathbb{E}_{(x,\tau) \sim \mathcal{D}_{cr}} \sum_{t \in \tau} \delta_{mask}(t) \log \pi_\theta(t\mid x, \tau_{:t}).
\end{equation}

\subsection{Direct Preference Optimization for Clarification-Rewriting Process}
\label{sec:process_dpo}

Previous studies have shown that LLMs exhibit phenomena of overthinking~\cite{overthinking} and underthinking~\cite{underthinking}. In the iterative clarification-rewriting framework we propose, the models need to avoid issues of overthinking and underthinking. On the one hand, the model should not ask overly detailed clarification questions that deviate from the original query's intent (i.e., overthinking). On the other hand, the model should not end asking clarification questions prematurely in cases of insufficient information (i.e. underthinking). In addition, in the previous data construction phase, we found that LLMs sometimes raise two clarification questions in one step. We expect the model to break down rewriting into individual sub-problems.

To address the above issues, we propose Direct Preference Optimization for Clarification-Rewriting Process (DPO-CRP), which constructs preference data from three perspectives: overthinking, underthinking, and insufficient decomposition. This enables the model to ask appropriate clarification questions at the right time. First, we take the data constructed in Section~\ref{sec:data_con} as chosen samples. Assuming the chosen sample's corresponding iterative trajectory is $\tau_{w} = (c_1,r_1,\dots,c_n,r_n)$. Then, rejected trajectories are constructed from three perspectives: overthinking, underthinking, and insufficient decomposition as follows.

\noindent \textbf{Overthinking}
To build data for overthinking, we need to ensure that the iterative process of the model has redundant steps. Therefore, we add an extra step $(c_{n+1}, r_{n+1})$ to the original iterative trajectory $\tau_{w}$, thereby obtaining the iterative trajectory of overthinking: $\tau_{ot} = (c_1,r_1,\dots,c_n,r_n,c_{n+1},r_{n+1})$. To ensure that this step is redundant, we verify whether $F(r_{n+1}) \leq F(r_n)$ during the sampling process:
\begin{equation}
\begin{split}
c_{n+1} \sim \mathcal{M}(r_{n}), r_{n+1} \sim \mathcal{M}(c_{n+1}, \mathcal{H}, r_{n}), \\s.t. F(r_{n+1}) \leq F(r_n).
\end{split}
\end{equation}
In this way, overthinking preference dataset $D_{ot} = \{x^{(i)},\tau_{w}^{(i)},\tau_{ot}^{(i)}\}$ is constructed.

\noindent \textbf{Underthinking}
We noticed that the underthinking iteration trajectory ends the iteration before reaching the optimal query. To construct such data, we randomly truncate the trajectory. Specifically, for the trajectory $\tau_{w}$, we randomly sample a position $e$ from $[1, n-1]$, then the corresponding underthinking trajectory is $\tau_{ut} = (c_1,r_1,\dots,c_e,r_e)$. Then the underthinking preference dataset is $D_{ut} = \{x^{(i)},\tau_{w}^{(i)},\tau_{ut}^{(i)}\}$.

\noindent \textbf{Insufficient decomposition}
During the previous data construction process, we found that the model would continuously raise two clarification questions in one iteration. To alleviate this insufficient decomposition issue, we sample a position $i$ from $[1, n-1]$, and merge $(c_i, r_i)$ with $(c_{i+1}, r_{i+1})$ to form $([c_j; c_{j+1}], r_{j+1})$, where $[\cdot; \cdot]$ denotes the concatenation operation. As a result, we obtain the corresponding insufficiently decomposed trajectory $\tau_{id} = (c_1,r_1,\dots,[c_j; c_{j+1}],r_{j+1},\dots)$ and the preference dataset $D_{id} = \{x^{(i)},\tau_{w}^{(i)},\tau_{id}^{(i)}\}$.

\noindent \textbf{Direct Preference Optimization} By constructing the rejected trajectories for the above three dimensions, we can obtain a process preference optimization dataset $D_{pref}=D_{ot} \cup D_{ut} \cup D_{id}$. Specific information about dataset $D_{pref}$ can be found in Appendix~\ref{sec:dataset}. After that, we can optimize the policy model $\pi_\theta$ by minimizing the following loss on dataset $D_{pref}$ through DPO~\cite{DPO} training:
\begin{equation}
\small{
    \begin{aligned}
        &\mathcal{L}_{\mathrm{DPO}}(\pi_\theta; \pi_{ref}; \mathcal{D}_{pref})
        =   -\mathbb{E}_{(x,\tau_w,\tau_l)\sim \mathcal{D}_{pref}} \\ & \bigg[ \log \sigma \bigg( \beta \log \frac{\pi_\theta(\tau_w | x)}{\pi_{\mathrm{ref}}(\tau_w | x)}
 - \beta \log \frac{\pi_\theta(\tau_l | x)}{\pi_{\mathrm{ref}}(\tau_l | x)} \bigg) \bigg],
    \label{eqn:prm-filter}
    \end{aligned}
}
\end{equation}
where $\pi_{\mathrm{ref}}$ is the reference model initialized from the original model before DPO.

\begin{table*}[t!]
    \centering
    \small
    \begin{threeparttable}
    \begin{tabular*}{1\textwidth}{clccccccccc}
        \toprule
         \multirow{2}{*}{{\textbf{Type}}} & \multirow{2}{*}{{\textbf{System}}} & \multirow{2}{*}{{\textbf{Backbone}}} & \multicolumn{4}{c}{\textbf{TopiOCQA}} &  \multicolumn{4}{c}{\textbf{QReCC}}  \\
         \cmidrule(lr){4-7} \cmidrule(lr){8-11}
         & & & \textbf{MRR} & \textbf{NDCG} & \textbf{R@10}  & \textbf{R@100} & \textbf{MRR} & \textbf{NDCG} & \textbf{R@10}  & \textbf{R@100} \\
        \midrule 
        \multirow{8}{*}{\rotatebox[origin=c]{90}{\textbf{Sparse (BM25)}}}
        & EDIRCS & T5-base & - & - & - & - & 41.2 & - & 62.7 & 90.2\\
        & LLM-Aided & ChatGPT & - & - & - & - & 49.4 & 46.5 & 67.1 & 88.2 \\
        & LLM4CS & ChatGPT & 27.9 & 26.4 & 48.4 & 71.1 & 51.6 & 49.3 & 75.3 & 92.6 \\
        & CHIQ & LLaMA2-7B & 25.6 & 23.5 & 44.7 & - & 54.3 & 51.9 & \textbf{78.5} & - \\
        & IterCQR & T5-base & 16.5 & 14.9 & 29.3 & 54.1 & 46.7 & 44.1 & 64.4 & 85.5 \\
        & AdaCQR & T5-base & 28.3 & 26.5 & 48.9 & 71.2 & 55.1 & 52.5 & 76.5 & 93.7 \\
        & RETPO & LLaMA2-7B  & 28.3 & 26.5 & 48.3 & 73.1 & 50.0 & 47.3 & 69.5 & 89.5 \\
        & \cellcolor{mygray}{ICR} & \cellcolor{mygray}LLaMA2-7B & \cellcolor{mygray}{\textbf{31.4}} & \cellcolor{mygray}{\textbf{30.4}} & \cellcolor{mygray}{\textbf{52.8}} & \cellcolor{mygray}{\textbf{76.3}} & \cellcolor{mygray}{\textbf{58.4}} & \cellcolor{mygray}{\textbf{54.9}} & \cellcolor{mygray}{78.3} & \cellcolor{mygray}{\textbf{95.9}} \\
        \midrule 
        \multirow{8}{*}{\rotatebox[origin=c]{90}{\textbf{Dense (ANCE)}}}
        & EDIRCS & T5-base & - & - & - & - & 42.1 & - & 65.6 & 85.3\\
        & LLM-Aided & ChatGPT & - & - & - & - & 43.5 & 41.3 & 65.6 & 82.3 \\
        & LLM4CS & ChatGPT & 35.4 & 34.4 & 55.2 & 72.2 & 44.7 & 41.8 & 67.2 & 84.0 \\
        & CHIQ & LLaMA2-7B & 38.0 & 37.0 & 61.6 & - & 47.2 & 44.6 & 70.8 & - \\
        & IterCQR & T5-base & 26.3 & 25.1 & 42.6 & 62.0 & 42.9 & 40.2 & 65.5 & 84.1 \\
        & AdaCQR & T5-base & 38.5 & 37.6 & 58.4 & 75.0 & 45.8 & 42.9 & 67.3 & 83.8 \\
        & RETPO & LLaMA2-7B  & 30.0 & 28.9 & 49.6 & 68.7 & 44.0 & 41.1 & 66.7 & 84.6 \\
        & \cellcolor{mygray}{ICR} & \cellcolor{mygray}LLaMA2-7B & \cellcolor{mygray}{\textbf{42.1}} & \cellcolor{mygray}{\textbf{40.4}} & \cellcolor{mygray}{\textbf{64.3}} & \cellcolor{mygray}{\textbf{78.3}} & \cellcolor{mygray}{\textbf{49.5}} & \cellcolor{mygray}{\textbf{46.8}} & \cellcolor{mygray}{\textbf{73.2}} & \cellcolor{mygray}{\textbf{88.3}}
        \\
        \bottomrule 
    \end{tabular*}
    \end{threeparttable}
    \caption{
    Evaluation results of various retrieval system types on the test sets of TopiOCQA and QReCC.
    }
    \vspace{-2mm}
    \label{table:main}
\end{table*}

\subsection{Process-aware Reciprocal Rank Fusion}
\label{sec:prrf}

During the inference phase, the model iteratively asks clarification questions and generates the rewritten queries based on the conversational context. We parse the model's output to obtain the set of rewritten queries $Q=\{r_1,r_2,\dots,r_{|Q|}\}$ in each iteration, where $r_i$ represents the rewritten query output by the model in the $i$-th iteration. In this process, multiple rewritten queries may be generated, each exhibiting distinct retrieval efficacy. Directly employing the query from the final iteration for retrieval can introduce noise due to error propagation throughout the clarification-rewriting iterative process, which may cause deviation from the original query. The primary challenge is effectively leveraging all generated rewritten queries.

We can notice that the retrieval performance of the rewritten queries corresponding to the later iterations is often better, which is due to the model's continuous refinement of the query during the clarification-rewriting process. Inspired by Reciprocal Rank Fusion~\cite{RRF}, we propose Process-Aware Reciprocal Rank Fusion (PRRF). Specifically, we weight the retrieval results corresponding to each rewritten query in the iterative process, with the weight coefficient being the position of the rewritten query in the iterations. The score of the passage $d$ after fusion is as follows:
\begin{equation}
\vspace{-2mm}
\label{eq:prrf}
    \texttt{PRRF}(d \in \mathcal{C}) =\sum\limits_{i \in [1,|Q|]} \frac{i}{rank(r_i, d) + k},
\end{equation}
where $rank(r_i, d)$ denotes the ranking of passage $d$ in the retrieval results of $r_i$ and $k$ is a constant that we set to 60. We multiply the weight of the rewritten query in the $i$-th iteration by $i$, so that the retrieval results of queries in later iterations have a greater impact on fusion.

\section{Experiments}

\subsection{Experimental Settings}
\label{sec:exp}
\noindent \textbf{Datasets and Evaluation Metrics} 
We conducted experiments on two widely used datasets, TopiOCQA~\cite{TopiOCQA} and QReCC~\cite{QReCC}. To evaluate the zero-shot capability of ICR, we also conducted zero-shot analysis on CAsT-19~\cite{CAsT-2019}, CAsT-20~\cite{CAsT-2020}, and CAsT-21~\cite{CAsT-2021}. Following previous work, we evaluated the performance of the models on four metrics: MRR, NDCG@3, Recall@10, and R@100. In Appendices~\ref{sec:dataset} and~\ref{sec:metrics}, we provide specific information on the datasets and evaluation metrics.

\noindent \textbf{Implementation Details}
In order to compare fairly with previous work, we use \texttt{gpt-3.5-turbo-0125} to generate iterative clarification-rewriting data and \texttt{Llama-2-7b-hf} as the backbone $\pi_\theta$. In data construction, the early stopping parameter $\mathcal{E}$ and the maximum number of iteration rounds $\mathcal{I}$ are set to 3 and 10, respectively. Following the previous work~\cite{CHIQ,AdaCQR,RETPO}, we also adopted query expansion. More implementation details can be found in Appendix~\ref{sec:implementation_details}.

\noindent \textbf{Baselines}
To verify the effectiveness of ICR, we compare it with the following baselines: EDIRCS ~\citep{EDIRCS}, LLM-Aided ~\citep{LLM-Aided}, LLM4CS ~\citep{LLM4CS}, IterCQR ~\citep{IterCQR}, CHIQ ~\citep{CHIQ}, AdaCQR ~\citep{AdaCQR} and RETPO ~\citep{RETPO}.

\subsection{Main Results}
As shown in Table~\ref{table:main}, ICR significantly outperforms all baselines in both sparse retrieval and dense retrieval. For example, in the dense retrieval of TopiOCQA, MRR, NDCG, R@10 and R@100 are improved by 3.6, 2.8, 5.9 and 3.3, respectively, compared to the best AdaCQR, indicating that ICR can not only recall more relevant passages but also rank them higher.
In comparison with the aforementioned LLMs-based baselines (e.g., CHIQ, IterCQR, AdaCQR and RETPO), ICR has made significant improvements. This finding indicates that the iterative process of clarification and rewriting enables the model to progressively refine the query, in contrast to the alternative of the model generating the rewritten query directly.

It is worth noting that both ICR and CHIQ used retrieval result fusion. However, CHIQ requires two rounds of query generation, while ICR can generate all rewritten queries at once. In addition, ICR considers the contributions of different rewritten queries for fusion. Therefore, ICR significantly outperforms CHIQ on almost all metrics, except for the R@10 metric on QReCC.

We also provide the performance analysis of ICR on CAsT 19-21 in the Appendix~\ref{sec:zero-shot}, which demonstrates that ICR has excellent zero-shot generalization capabilities.

\begin{table}[t]
    \small
    \centering
    \resizebox{\linewidth}{!}{
    \begin{tabular}{lcccc}
        \toprule
        & \multicolumn{4}{c}{\textbf{TopiOCQA}} \\
         \cmidrule(lr){2-5}
        \multicolumn{1}{c}{\textbf{Variant}} &  \textbf{MRR} & \textbf{NDCG} & \textbf{R@10}  & \textbf{R@100} \\        
        \midrule
        \cellcolor[gray]{0.9} \textit{ICR} & \cellcolor[gray]{0.9} 42.1 & \cellcolor[gray]{0.9} 40.4 & \cellcolor[gray]{0.9} 64.3  & \cellcolor[gray]{0.9} 78.3 \\

        \quad w/o. OT & 41.2$_{\downarrow 0.9}$ & 39.0$_{\downarrow 1.4}$ & 63.1$_{\downarrow 1.2}$ & 77.1$_{\downarrow 1.2}$ \\
        \quad w/o. UT & 40.9$_{\downarrow 1.2}$ & 38.9$_{\downarrow 1.5}$ & 62.9$_{\downarrow 1.4}$ & 77.2$_{\downarrow 1.1}$ \\
        \quad w/o. ID & 41.6$_{\downarrow 0.5}$ & 39.5$_{\downarrow 0.9}$ & 63.5$_{\downarrow 0.8}$ & 77.6$_{\downarrow 0.7}$ \\
        \quad w/o. DPO-CRP & 39.4$_{\downarrow 2.7}$ & 37.5$_{\downarrow 2.9}$ & 61.8$_{\downarrow 2.5}$ & 76.2$_{\downarrow 2.1}$ \\
        \midrule
        Vanilla SFT  & 41.6$_{\downarrow 0.5}$ & 40.1$_{\downarrow 0.3}$ & 63.7$_{\downarrow 0.6}$  & 77.8$_{\downarrow 0.5}$ \\
        Final Rewriting  & 41.3$_{\downarrow 0.8}$ & 39.2$_{\downarrow 1.2}$ & 63.4$_{\downarrow 0.9}$ & 77.4$_{\downarrow 0.9}$ \\
        RRF  & 40.9$_{\downarrow 1.2}$ & 38.7$_{\downarrow 1.7}$ & 63.1$_{\downarrow 1.2}$  & 76.9$_{\downarrow 1.4}$  \\
       
        ICR$_\texttt{GPT-4.1}$  & 42.8$_{\uparrow 0.7}$ & 41.1$_{\uparrow 0.7}$ & 64.9$_{\uparrow 0.6}$ & 78.7$_{\uparrow 0.4}$  \\
        ICR$_\texttt{Qwen3-14B}$  & 42.6$_{\uparrow 0.5}$ & 40.9$_{\uparrow 0.5}$ & 64.7$_{\uparrow 0.4}$ & 78.5$_{\uparrow 0.2}$  \\
        
        \bottomrule
    \end{tabular}
    }
    \caption{
    Ablation study for each component of \textsc{ICR} and different LLMs.
    }
    \label{table:ablation}
    \vspace{-6mm}
\end{table}

\subsection{Ablation Study}
To analyze the contribution of each component, we performed ablation experiments shown in Table~\ref{table:ablation}.

\noindent \textbf{DPO-CRP} To optimize the clarification-rewriting iteration process, we designed preference data from three perspectives: overthinking, underthinking, and insufficient decomposition. We have separately removed the preference data for the above three dimensions, as shown in ``w/o OT'' (OverThinking), ``w/o UT'' (UnderThinking), and ``w/o ID'' (Insufficient Decomposition) in Table~\ref{table:ablation}, respectively. Removing any preference dimension reduces retrieval performance, showing that all three dimensions are crucial for effective clarification-rewriting iterations. Notably, eliminating the underthinking preference data causes the most significant drop in performance, potentially due to the fact that the model is more inclined to end iterations prematurely as opposed to iterating redundant steps. We also tried to remove the whole DPO-CRP (``w/o DPO-CRP'') directly. If only supervised fine-tuning is performed without preference optimization, the model's performance has dropped significantly.

\noindent \textbf{PSFT} Recognizing the distinct skills of asking clarification questions and rewriting, we propose Progressive Supervised Fine-Tuning (PSFT). We also tried directly fine-tuning the model for 3 epochs without masking any tokens (``Vanilla SFT''), and the retrieval performance has declined across all metrics. This indicates that PSFT can guide the model from asking clarification questions to rewriting.

\noindent \textbf{PRRF} To analyze the effectiveness of our proposed Process-aware Reciprocal Rank Fusion (PRRF), we evaluated two additional variants, i.e., directly using the rewritten query from the last iteration (``Final rewriting'') and reciprocal rank fusion without considering the contribution of rewritten queries (``RRF''). It can be observed that using the final rewritten query and using the reciprocal rank fusion both lead to performance degradation. Additionally, the retrieval performance of RRF is worse than that of directly using the final rewritten query. This indicates that some queries in the early iteration process are still incomplete, and their retrieval results may interfere with the fusion process.

\noindent \textbf{Different LLMs} To analyze whether ICR can be adapted to other models, we used \texttt{GPT-4.1} for data construction and \texttt{Qwen3-14B} as the backbone for training, with results shown in Table~\ref{table:ablation} for ICR$_\texttt{GPT-4.1}$ and ICR$_\texttt{Qwen3-14B}$. It can be observed that using larger and more advanced models can achieve better performance, which indicates that ICR can be adapted to other models. Notably, ICR$_\texttt{GPT-4.1}$ exhibits a more pronounced performance enhancement relative to ICR$_\texttt{Qwen3-14B}$, potentially attributable to the critical role of high-quality iterative trajectory data over employing a robust backbone model.

\begin{figure}[t]
\begin{center}
 \includegraphics[width=0.7\linewidth]{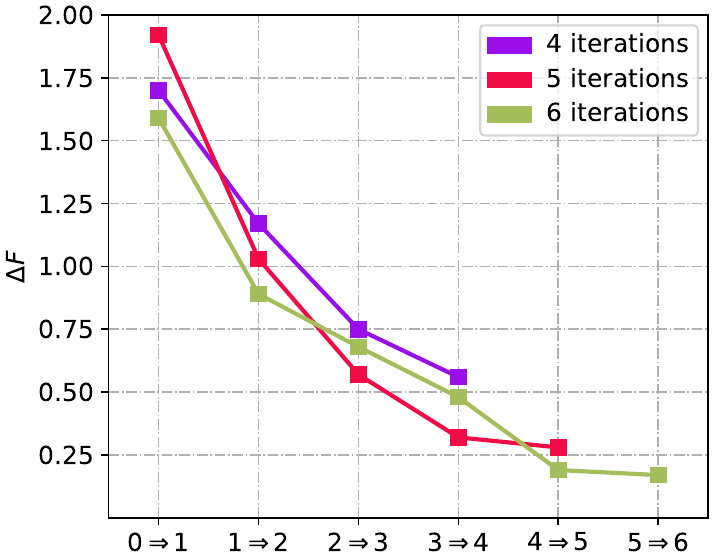}
 \caption{ The magnitude of change $\Delta F$ in the retrieval performance of queries between adjacent iteration steps.}
 \label{fig:adjacent_change}
\end{center}
\vspace{-0.7cm}
\end{figure}

\subsection{Analysis}
\noindent \textbf{Evolution of Queries between Adjacent Iterative Steps} We analyze the variation of retrieval performance of queries between adjacent iteration steps for samples with the iteration steps 4, 5, and 6, as shown in Figure ~\ref{fig:adjacent_change}. We can observe that in the initial iterations, the growth of retrieval performance (i.e. $F(\cdot)$) is very rapid. As the iterations increase, the growth rate gradually slows down. This phenomenon arises from the saturation of query information, whereby additional clarification becomes redundant and potentially detrimental to retrieval performance.

\begin{table}[t]
    \centering

\resizebox{0.82\linewidth}{!}{
\begin{tabular}{ccccc}
\specialrule{1pt}{1pt}{1pt}
\multicolumn{1}{c|}{}                                          & \multicolumn{2}{c}{\textbf{TopiOCQA}} & \multicolumn{2}{c}{\textbf{QReCC}} \\
\multicolumn{1}{c|}{\textbf{}}                                 & Sparse  & Dense & Sparse     & Dense      \\ \midrule
\multicolumn{5}{c}{\cellcolor[HTML]{9dbcd4}\textit{\textbf{Before DPO-CRP}}}                                                                        \\ \midrule
\multicolumn{1}{c|}{\texttt{LSR}}    & 77.8             & 78.4             & 79.8          & 79.1            \\
\multicolumn{1}{c|}{\texttt{GSR}} & 60.2             & 60.4             & 62.4          & 61.2            \\ \midrule
\multicolumn{5}{c}{\cellcolor[HTML]{c2ff89}\textit{\textbf{After DPO-CRP}}}                                                                                 \\ \midrule
\multicolumn{1}{c|}{\texttt{LSR}}    & 80.2$_{\uparrow 2.4}$             & 81.0$_{\uparrow 2.6}$             & 83.2$_{\uparrow 3.4}$          & 82.8$_{\uparrow 3.7}$            \\
\multicolumn{1}{c|}{\texttt{GSR}}  & 63.2$_{\uparrow 3.0}$    & 64.2$_{\uparrow 3.8}$    & 66.4$_{\uparrow 4.0}$ & 65.7$_{\uparrow 4.5}$   \\\bottomrule[1pt]
\end{tabular}
}
    \caption{\texttt{LSR} and \texttt{GSR} on TopiOCQA and QReCC.}
    \label{table:slg}
    \vspace{-0.4cm}
\end{table}

\noindent \textbf{Analysis of Iterative Process} The iterative nature of ICR can lead to error propagation, where inappropriate clarification questions in earlier iterations result in a rewritten query that contradicts the original intent. To quantitatively assess the iterative improvement of query quality, we define the local success rate \texttt{LSR} and the global success rate \texttt{GSR}:
\begin{equation*}
\label{eq:S_g}
\begin{aligned}
    &\resizebox{.95\hsize}{!}{$\texttt{LSR}=\frac{1}{N}\sum_{i=1}^{N}  \frac{1}{|\mathcal{D}_{test}^{i}|}\sum_{r_i^j\in\mathcal{D}_{test}^{i}}\mathbbm{1}  [(F(r_i^{j-1}) <  F(r_i^j) ) ]$,} \\
    &\resizebox{.9\hsize}{!}{$\texttt{GSR}=\frac{1}{N}\sum_{i=1}^{N} \mathbbm{1} \left [ \bigwedge_{r_i^j\in\mathcal{D}_{test}^{i}}(F(r_i^{j-1}) <  F(r_i^j) )\right ]$,}
\end{aligned}
\end{equation*}

\noindent where $\mathbbm{1}\left ( \cdot \right )$ is the indicator function, and $N$ is the number of data entries. $r_i^j$ represents the rewritten query corresponding to the $j$-th iterative step of the $i$-th test sample $\mathcal{D}_{test}^{i}$ (specifically, $r_i^0$ represents the original query). $F(\cdot)$ is the metric used to calculate the quality of the rewritten query in Equation~\ref{eq:rs_calculation}. \texttt{LSR} and \texttt{GSR} measure the proportion of effective iterative steps, i.e., the clarification questions posed by the model during the iteration process enhance the retrieval performance of the rewritten query. \texttt{LSR} and \texttt{GSR} are measured from both step-level and sample-level respectively. As shown in Table~\ref{table:slg}, both \texttt{LSR} and \texttt{GSR} are significantly improved after using DPO-CRP. This indicates that the preference samples designed in three dimensions can effectively guide the clarification and rewriting process. Moreover, We observed that the improvement in \texttt{GSR} is greater than \texttt{LSR} after using DPO-CRP. This is because training with DPO-CRP corrected minor errors caused by only a few steps of the iteration trajectory in certain samples, thereby increasing the proportion of effective trajectories.

We also analyze the number of redundant steps in the overthinking preference data in Appendices~\ref{sec:pd_con}, and provide a latency analysis of ICR in Appendix~\ref{sec:latency}.

\begin{table}[t]
    \centering
    \small
\setlength{\tabcolsep}{3pt}
    \begin{tabular}{cccccc}
        \toprule
         \multirow{2}{*}{{$\mathbf{F(\cdot)}$}} & \multirow{2}{*}{{\textbf{Type}}} & \multicolumn{4}{c}{\textbf{TopiOCQA}}  \\
         \cmidrule(lr){3-6}
         & & \textbf{MRR} & \textbf{NDCG} & \textbf{R@10}  & \textbf{R@100} \\
        \midrule 
        \multirow{2}{*}{{$m_s(\cdot)+m_d(\cdot)$}} & Sparse
        & 31.4 & 30.4 & 52.8 & 76.3  \\
        & Dense
        & 42.1 & 40.4 & 64.3 & 78.3  \\
        \midrule
        \multirow{2}{*}{{$m_s(\cdot)$}} & Sparse
        & 29.3 & 28.5 & 50.8 & 74.2  \\
        & Dense
        & 39.8 & 37.8 & 61.7 & 75.9  \\
        \midrule 
        \multirow{2}{*}{{$m_d(\cdot)$}} & Sparse
        & 28.7 & 27.8 & 49.9 & 73.7 \\
        & Dense
        & 40.1 & 38.2 & 62.2 & 76.6 \\
        \bottomrule 
    \end{tabular}
    \caption{
    Impact on model performance when different retrieval types are used as query quality indicators.
    }
    \vspace{-2mm}
    \label{table:rq_pm}
\end{table}

\subsection{Quality Measurement of Rewritten Queries}
\label{sec:rq_pm}
During the process of constructing iterative data, we measure the quality of queries by combining the performance of sparse retrieval and dense retrieval. To verify the impact of different types of retrieval performance on measuring query quality, we attempted using only sparse retrieval and only dense retrieval separately, as shown in Table~\ref{table:rq_pm}. It can be observed that using only sparse retrieval (i.e., $m_s(\cdot)$) or only dense retrieval (i.e., $m_d(\cdot)$) to measure query quality leads to degradation of model performance. In addition, when using sparse retrieval to measure query quality, the performance of dense retrieval decreases more significantly, while when using dense retrieval to measure query quality, the performance of sparse retrieval decreases more significantly. This is due to the fact that sparse and dense retrieval take into account lexical and semantic level information respectively, and measuring performance using only one type of retrieval will lead to bias.

\subsection{Correlation between Iterations and Performance}
\label{sec:iter_analyze}

\begin{figure}[t]
\begin{center}
 \includegraphics[width=0.9\linewidth]{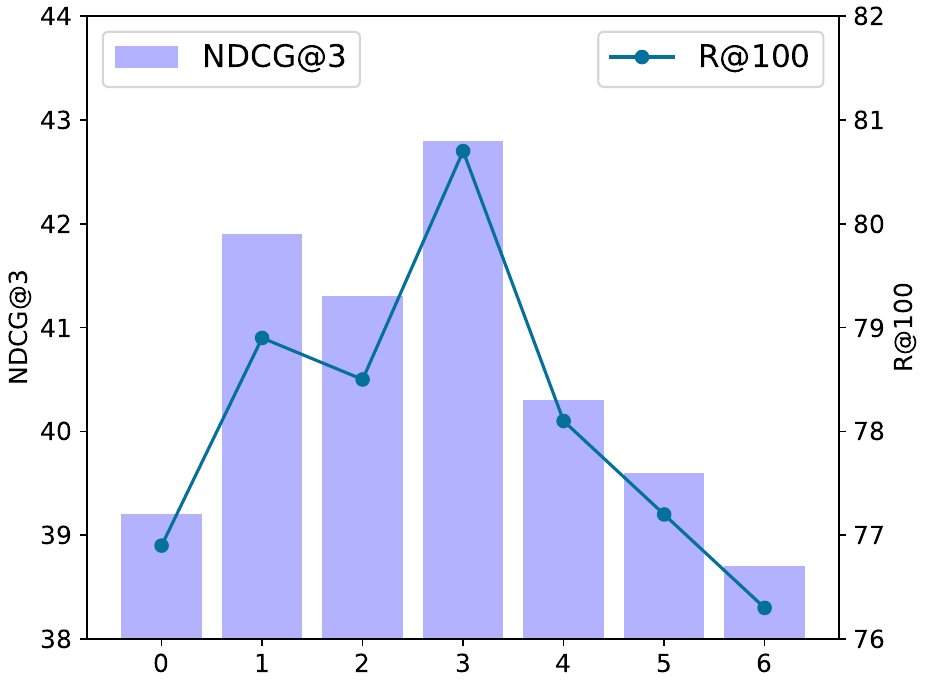}
 \caption{The correlation between iterative steps and model performance.}
 \label{fig:step2mt}
\end{center}
 \vspace{-0.7cm}
\end{figure}

\begin{figure}[t]
\begin{center}
 \includegraphics[width=0.9\linewidth]{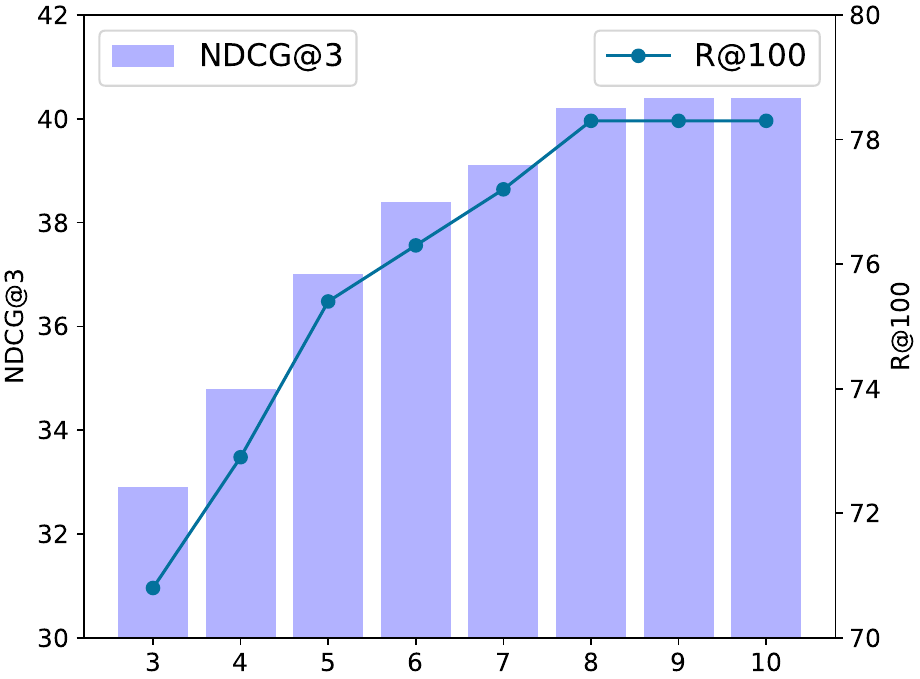}
 \caption{Impact of maximum number of iteration rounds $\mathcal{I}$ on model performance.}
 \label{fig:iter_num}
\end{center}
 \vspace{-0.7cm}
\end{figure}

The number of iterations in the clarification and rewriting process varies depending on the query. As shown in Figure~\ref{fig:step2mt}, we analyzed the correlation between the number of iterations and model performance on the TopiOCQA dataset. It can be observed that the performance is best when the number of iteration steps is 3. This could be because of insufficient rewriting with few iterations. On the other hand, excessive iterations can lead to unnecessary clarification questions that introduce irrelevant noise in the query, which hinders retrieval performance. This validates the rationality of the overthinking and underthinking data we designed in the DPO-CRP process.

\subsection{Maximum Number of Iterations $\mathcal{I}$ in Data Construction}
\label{sec:max_iter}

During the iterative data construction phase, we default to a maximum of 10 iterations, exiting when the iteration number reaches 10. We analyzed the impact of the maximum number of iterations $\mathcal{I}$ on the model performance shown in Table~\ref{fig:iter_num}. When the number of iterations is less than 8, increasing the number of iterations continuously improves the model performance. This is due to the fact that when the maximum number of iterations is small, the model may exit the iteration prematurely, resulting in insufficient information in the rewritten query. This situation leads to bias in the generated trajectory data, thereby preventing the model from correctly iterating.

\subsection{Qualitative Analysis and Human Evaluation}
To qualitatively assess the evolution of queries during the clarification-rewriting iteration process, we define two metrics: \emph{Completeness} and \emph{Factuality}. \emph{Completeness} assesses whether a query encapsulates all necessary information, while \emph{Factuality} evaluates the alignment of query content with the dialogue context. It is obvious that during the iteration process, the \emph{Completeness} of the query does not decrease, while its \emph{Factuality} does not increase. This phenomenon occurs because while information is continuously refined, there is also potential for noise introduction.

We sampled 200 instances with more than one iteration step and distributed these samples to three graduate students in NLP. They annotated the changes in \emph{Completeness} and \emph{Factuality} of the queries from the initial to the final iteration, as shown in Table~\ref{tab:comp_fact}. It can be observed that the \emph{Completeness} and \emph{Factuality} of most queries improved in the final iteration, accounting for 61.0\%. Additionally, the retrieval performance of these queries has significantly improved ($\uparrow$4.10). For 28.0\% of the queries, \emph{Completeness} and \emph{Factuality} remained constant due to either no alterations in subsequent iterations or modifications limited to stop words, resulting in unchanged retrieval performance. 

During the iterations, although in some cases the model rewrite the query to be more complete based on the clarification questions, it also introduces noise, which account for 9.0\% and result in an average performance degradation of 1.32. We note that 2.0\% of the queries have unchanged \emph{Completeness} but degraded \emph{Factuality}, which is due to the rewriting of correct information into incorrect information in later iterations, resulting in degraded \emph{Factuality} and reduced retrieval performance.

\begin{table}
\centering
\begin{tabular}{ccccc}
\hline
\textbf{Comp.} & \textbf{Fact.} & Proportion (\%) & Avg. $\Delta F$ \\
\hline
\textcolor{red}{\ding{55}} & \textcolor{red}{\ding{55}} & 2.0 & -3.61\\
\textcolor{green}{\ding{51}} & \textcolor{red}{\ding{55}} & 9.0 & -1.32\\
\textcolor{red}{\ding{55}} & \textcolor{green}{\ding{51}} & 28.0 & 0 \\
\textcolor{green}{\ding{51}} & \textcolor{green}{\ding{51}} & 61.0 & 4.10\\
\hline
\end{tabular}
\caption{\label{tab:comp_fact}
Analysis of completeness and factuality changes during the iterative process. \textcolor{red}{\ding{55}} indicates unchanged completeness (Comp.) or decreased factuality (Fact.), while \textcolor{green}{\ding{51}} indicates increased completeness or unchanged factuality.
}
\vspace{-0.4cm}
\end{table}

\subsection{Case Study}
\begin{figure}[t]
\begin{center}
 \includegraphics[width=1\linewidth]{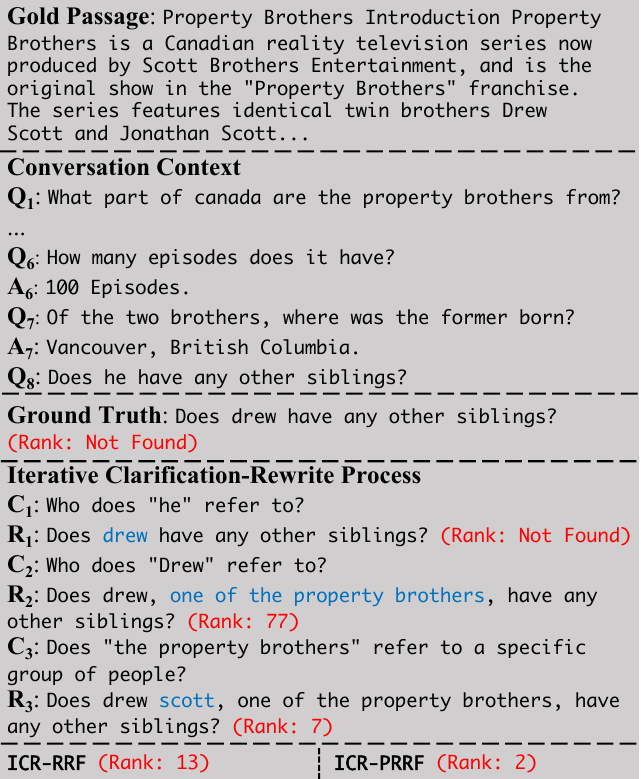}
 \caption{Case study on TopiOCQA. The red font indicates the ranking of the gold passage in the top-100 retrieval results.}
 \label{fig:case}
\end{center}
\vspace{-0.7cm}
\end{figure}
In Figure~\ref{fig:case}, we provide a case study where the blue font indicates the newly added key information for retrieval in each iteration. In this example, the model undergoes three clarification-rewriting iterations. The rewritten query generated in each iteration introduced key information that was not present in the previous query. In the first iteration, the model rewrites ``he'' as ``drew''. In the second iteration, the model provided an explanation of the background of ``drew'' (i.e., ``one of the property brothers''). By inferring from ``one of the property brothers'', the model introduced ``scott'' in the third iteration, making the name more specific. During the retrieval process, we found that the effect of Reciprocal Rank Fusion (ICR-RRF) was worse than directly using the final rewritten query (13 vs. 7), which is attributed to the interference of rewritten queries in the first two iterations on the fusion. By using Process-aware Reciprocal Rank Fusion (ICR-PRRF), the weights of the queries in the subsequent iterations can be improved, ultimately ranking the gold passage second in the top-100 retrieval results. In addition, ICR makes query rewriting more transparent by pivoting on clarification questions. We provide more examples in Appendix~\ref{sec:examples}.

\section{Conclusion}
In this paper, we incorporate a novel framework ICR into CQR. ICR involves progressive fine-tuning and constructing preference data for preference alignment from three perspectives: overthinking, underthinking, and insufficient decomposition. Through process-aware fusion, all rewritten queries in the iterative process are fully utilized. On both TopiOCQA and QReCC datasets, our ICR achieves state-of-the-art performance.

\section*{Limitations}
Although our proposed ICR can advance the research on conversational search, it still suffers from the following two drawbacks. First, we used rewritten queries for the final retrieval, but did not utilize the clarification questions. However, clarification questions often contain some key information. Future research can focus on achieving more efficient retrieval by combining clarification questions with rewritten queries. Second, in the retrieval result fusion module, we fused all the rewritten queries. Even if we consider the contributions of different rewritten queries, some rewritten queries may deviate significantly from the original query and can be directly discarded.

\section*{Acknowledgements}
The authors would like to thank the three anonymous reviewers for their comments on this paper. This research was supported by the National Natural Science Foundation of China (Nos. 62376181 and 62276177), and Project Funded by the Priority Academic Program Development of Jiangsu Higher Education Institutions.

\bibliography{custom}
 \newpage
\appendix

\begin{table}[h]
\centering
\small
\setlength{\tabcolsep}{4pt}{
\begin{tabular}{llrrr}
\toprule
Dataset & Split & \#Conv. & \#Turns(Qry.) & \#Collection \\ \midrule
\multirow{2}{*}{TopiOCQA} & Train & 3,509 & 45,450 & \multirow{2}{*}{25M} \\
 & Test  & 205 & 2,514 & \\
\midrule
\multirow{2}{*}{QReCC} & Train & 10,823 & 63,501 & \multirow{2}{*}{54M} \\
 & Test  & 2,775 & 16,451 & \\
\midrule
{CAsT-19} & Test  & 50 & 479 & \multirow{2}{*}{{38M}} \\
{CAsT-20} & Test  & 25 & 208 & \\ 
\midrule
{CAsT-21} & Test  & 26 & 239 & 40M \\ 
\bottomrule
\end{tabular}}
\caption{Statistics of conversational search datasets.}
\vspace{-2ex}
\label{table:datasets}
\end{table}

\begin{table}[h]
\centering
\small
\setlength{\tabcolsep}{7pt}{
\begin{tabular}{lcccc}
\toprule
Dataset & \#$D_{ot}$ & \#$D_{ut}$ & \#$D_{id}$ & \#$D_{pref}$ \\ \midrule
\multirow{1}{*}{TopiOCQA} & 45,450 & 39,112 & 39,112 & 123,674 \\
\midrule
\multirow{1}{*}{QReCC} & 29,596 & 26,427 & 26,427 & 82,450 \\
\bottomrule
\end{tabular}}
\caption{Statistics of the process preference optimization dataset.}
\vspace{-2ex}
\label{table:dpo_datasets}
\end{table}

\section{Prompts}
\label{sec:prompt}
In Tables~\ref{tab:clarification_prompt} and~\ref{tab:rewriting_prompt}, we provide prompts for generating clarification questions and rewriting queries based on clarification questions, respectively.

\section{Details of Datasets}
\label{sec:dataset}
We present statistical information on the datasets used in the experiments in Table~\ref{table:datasets}. It is worth noting that for the QReCC dataset, following previous work~\cite{CHIQ,AdaCQR,RETPO}, some samples without gold passage label were excluded, resulting in 29,596 and 8,209 turns being retained in the final training set and test set, 
 respectively.

For the construction of the clarification-rewriting process preference optimization dataset $D_{pref}$, we devised three dimensions: overthinking, underthinking, and insufficient decomposition. Among them, for underthinking and insufficient decomposition, we filter out iterative trajectories with rounds less than 2 because these trajectories cannot be further truncated or merge the intermediate steps. The statistics of the final process preference optimization dataset $D_{pref}$ are shown in Table~\ref{table:dpo_datasets}.

\begin{figure}[h]
\begin{center}
 \includegraphics[width=1\linewidth]{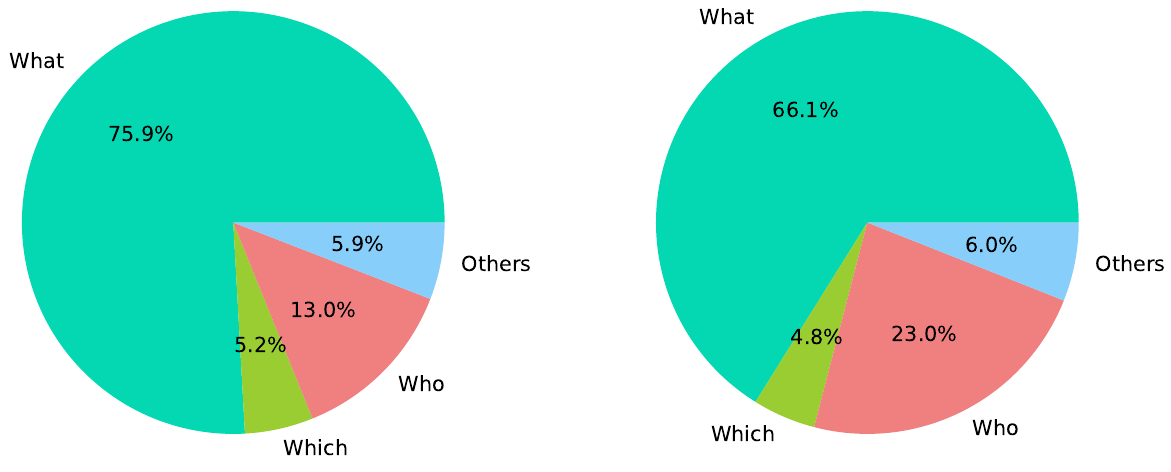}
 \caption{Distribution of number of interrogative words in clarification questions on TopiOCQA (left) and QReCC (right).}
 \label{fig:cl_num}
\end{center}
\vspace{-0.4cm}
\end{figure}

In order to analyze the distribution of types of clarification questions, we counted the number of interrogative words in the clarification questions, as shown in Figure~\ref{fig:cl_num}. The interrogative words with a percentage greater than 1\% are ``What'', ``Which'', and ``Who''. The interrogative words included ``Others'' are ``How'', ``Where'', ``When'', and so on. The interrogative word "What" has the highest proportion in clarification questions, followed by ``Who''. These two interrogative words usually involve referring to some entities.

\section{Details of Evaluation Metrics}
\label{sec:metrics}
MRR focuses on the position at which relevant passages are ranked, with a higher MRR indicating a higher position. NDCG@3 considers the relevance and ranking position of the top 3 retrieval results. Recall@K evaluates the ability to retrieve relevant passages in the top-K results.

\section{Implementation Details}
\label{sec:implementation_details}
We used the LoRA~\citep{lora} fine-tuning method, with the LoRA rank set to 8. The model performs DPO training of 3 epochs on the constructed process preference dataset. We set the following hyperparameters for optimization: batch size is 8, gradient accumulation steps are 4, and learning rate is 1e-5. For the retrieval system, we use Faiss~\cite{Faiss} for dense retrieval and Pyserini~\cite{Pyserini} for sparse retrieval. In BM25, $k_1$ and $b$ are set to 0.9 and 0.4 on TopiOCQA, and 0.82 and 0.68 on QReCC, where $k_1$ controls the non-linear term frequency normalization, while $b$ is the scale of the inverse document frequency. The retriever used in dense retrieval is ANCE~\cite{ANCE}. We use \texttt{pytrec\_eval} to calculate the evaluation metrics MRR, NDCG@3, Recall@10, and R@100.

\begin{table*}[ht]
\centering
\resizebox{\linewidth}{!}{
\begin{tabular}{lcccccccccc}
\toprule
\multicolumn{1}{l}{\multirow{2}{*}{System}} & \multirow{2}{*}{Backbone} & \multicolumn{3}{c}{CAsT-19} & \multicolumn{3}{c}{CAsT-20} & \multicolumn{3}{c}{CAsT-21} \\ 
\cmidrule(lr){3-5} \cmidrule(lr){6-8} \cmidrule(lr){9-11} 
& & MRR & N@3  & R@10  & MRR & N@3 & R@10 & MRR     & N@3 & R@10  \\ 
\midrule
E5-Mistral & Mistral-7B & 62.2 & 31.3 & 9.5 & 22.0 & 15.4 & 8.4 & 48.2 & 32.5 & 20.5\\
HyDE & ChatGPT-3.5 & 55.6 & 39.2 & 10.0 & 44.8 & 29.3 & 16.9 & - & - & -\\
Query2doc & ChatGPT-3.5 & 58.8 & 42.4 & 11.6 & 48.6 & 32.5 & 17.3 & - & - & -\\
InstructorR & ChatGPT-3.5 & 61.2 & 46.6 & 10.4 & 43.7 & 29.6 & 8.3 & 46.7 & 32.5 & 18.4 \\
LLM4CS & ChatGPT-3.5 & 70.4 & 46.8 & 11.7 & 58.6 & 41.5 & 19.3 & 66.1 & 46.9 & 24.4\\
RepLLaMA & LLaMA2-7B & 62.4 & 31.6 & 10.6 & 26.8 & 18.3 & 10.4 & 47.4 & 32.7 & 19.6 \\
LLM-Embedder & LLaMA2-7B & 63.3 & 36.6 & 11.4 & 25.2 & 15.4 & 8.7 & 46.8 & 31.2 & 17.3\\
\fname{} &  LLaMA2-7B & 73.3 & 50.5 & 12.9 & 54.0 & 38.0 & 19.3 & 62.9 & 46.5 & 25.2 \\
AdaCQR &  T5-base & 74.5 & - & 13.8 & 56.6 & - & 19.2 & 64.2 & - & 25.0 \\
\midrule
ICR &  LLaMA2-7B & \textbf{76.9} & \textbf{53.7} & \textbf{15.3} & \textbf{58.9} & \textbf{43.8} & \textbf{20.3} & \textbf{67.9} & \textbf{47.8} & \textbf{27.1} \\

        \bottomrule
     \end{tabular}}
     \caption{Zero-shot retrieval performances under the dense retrieval (ANCE).}
     \label{tab:zero-shot}
\end{table*}

\section{Zero-shot Analysis}
\label{sec:zero-shot}
To analyze the zero-shot generalization performance of ICR, we additionally compared ICR with the following methods: LLM-Embedder \cite{LLM-Embedder}, HyDE \cite{HyDE}, Query2doc \cite{Query2doc}, InstructorR \cite{InstructorR}, RepLLaMA \cite{RepLLaMA} and E5-Mistral \cite{E5-Mistral}.

We report in Table~\ref{tab:zero-shot} the generalization performance of ICR on three datasets: CAST-19, CAST-20, and CAST-21. ICR significantly outperforms previous methods in zero-shot setting, with improvements of 2.4, 2.3, and 3.7 on the MRR of three datasets compared to previous methods. This is attributed to the fact that ICR's iterative rewriting framework can continuously refine the query, propose clarification questions on the ambiguous expressions of the query, and perform rewriting.

\begin{table}[t]
    \small
    \centering
    \resizebox{\linewidth}{!}{
    \begin{tabular}{lcccc}
        \toprule
        & \multicolumn{4}{c}{\textbf{TopiOCQA}} \\
         \cmidrule(lr){2-5}
        \multicolumn{1}{c}{\textbf{Variant}} &  \textbf{MRR} & \textbf{NDCG} & \textbf{R@10}  & \textbf{R@100} \\        
        \midrule
        \cellcolor[gray]{0.9} \textit{ICR} & \cellcolor[gray]{0.9} 42.1 & \cellcolor[gray]{0.9} 40.4 & \cellcolor[gray]{0.9} 64.3  & \cellcolor[gray]{0.9} 78.3 \\
        \midrule
        Multi-Redundant  & 42.3 & 40.6 & 63.8  & 77.8 \\
        \bottomrule
    \end{tabular}
    }
    \caption{
    Analysis of the number of redundant iterative steps in the process of constructing preference data for overthinking.
    }
    \label{table:sample_con}
    \vspace{-3mm}
\end{table}

\section{Analysis of Overthinking Preference Data Construction}
\label{sec:pd_con}
When constructing overthinking data, we build the trajectory of overthinking by adding a redundant step after the iterative trajectory of the chosen sample. We can also choose to add more redundant steps. Specifically, we randomly sample a number $k$ from $[1, 2, 3, 4]$, and add $k$ redundant steps after $\tau_{w} = (c_1,r_1,\dots,c_n,r_n)$, ensuring $F(r_{n}) \geq F(r_{n+1}) \geq \dots \geq F(r_{n+k})$. As shown in ``Multi-Redundant'' of Table~\ref{table:sample_con}, increasing the number of redundant iterative steps does not significantly improve performance, and even results in a decrease in performance at R@10 and R@100. One potential reason is that the iterative trajectory after adding more redundant steps differs too much from the original iterative trajectory, widening the gap between chosen samples and rejected samples. The excessive distinction may cause the model to overfit the training data during the preference learning process, unable to grasp the timing to exit the iteration.

\section{Latency Analysis of ICR}
\label{sec:latency}
\begin{table}[t]
\small
    \centering
 \setlength{\tabcolsep}{18pt}
\begin{tabular}{ccc}
\specialrule{1pt}{1pt}{1pt}
\multicolumn{1}{c}{}                                          & \multicolumn{1}{c}{\textbf{TopiOCQA}} & \multicolumn{1}{c}{\textbf{QReCC}} \\ \midrule
\multicolumn{1}{c}{Latency}    & \multicolumn{1}{c}{2.13}             & \multicolumn{1}{c}{2.17}            \\
\bottomrule[1pt]
\end{tabular}

    \caption{Per-sample latency (in seconds) for iterative trajectory generation on TopiOCQA and QReCC.}
    \label{table:latency}
\end{table}

Since ICR requires the generation of complete iterative trajectories during inference, it may incur additional computational overhead compared to directly generating rewritten query. We calculated the latency of ICR from iterative trajectory generation, as shown in Table~\ref{table:latency}. The inference latency on TopiOCQA and QReCC is 2.13 seconds and 2.17 seconds respectively, which is tolerable for humans and can be deployed in the real world.

\section{Examples of ICR}
\label{sec:examples}
\begin{figure}[t]
\begin{center}
 \includegraphics[width=1\linewidth]{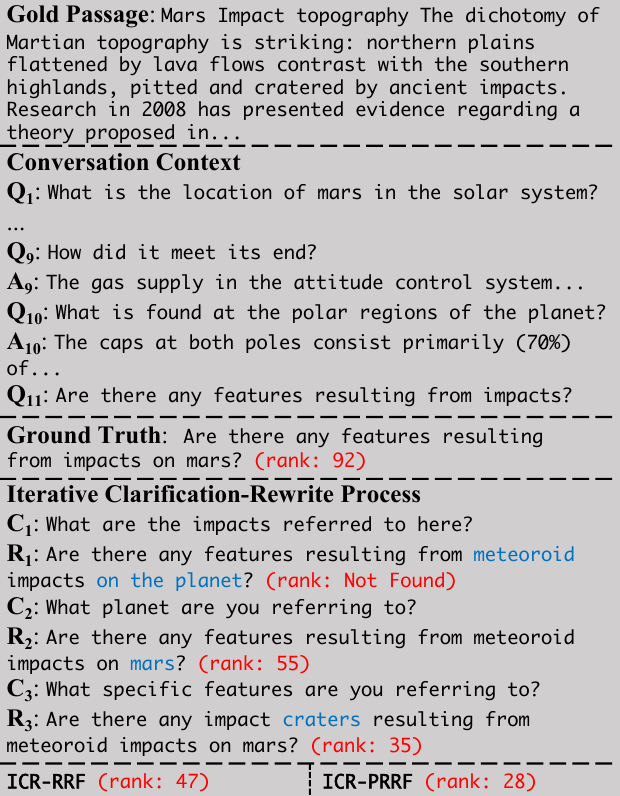}
 \caption{Example on TopiOCQA.}
 \label{fig:case1}
\end{center}
\vspace{-0.2cm}
\end{figure}

\begin{figure}[t]
\begin{center}
 \includegraphics[width=1\linewidth]{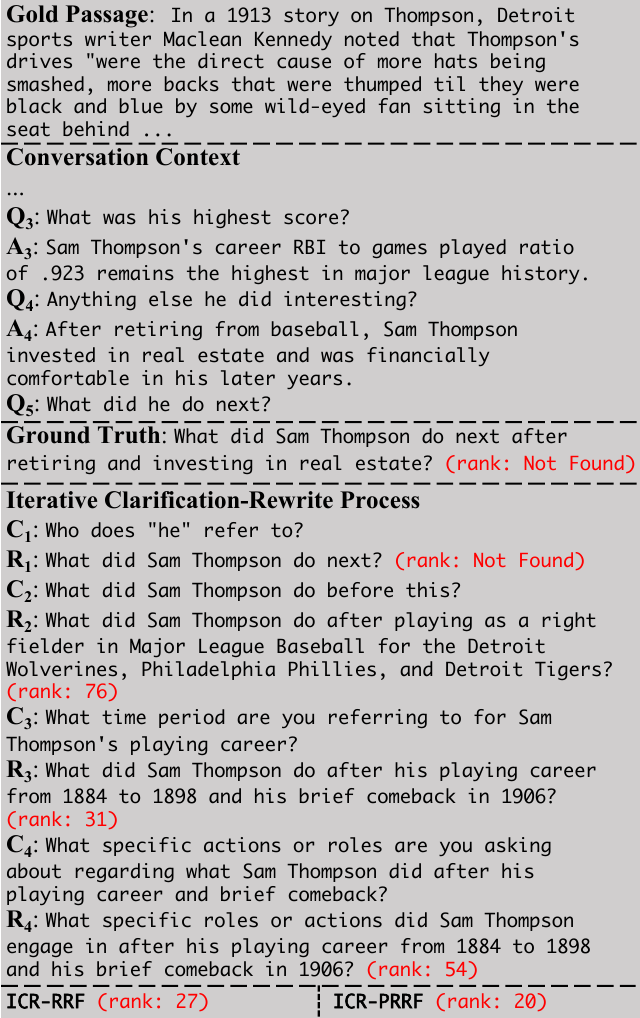}
 \caption{Example on QReCC.}
 \label{fig:case2}
\end{center}
\vspace{-0.3cm}
\end{figure}

In addition to the example in Figure~\ref{fig:case}, we provide two examples in Figures~\ref{fig:case1} and~\ref{fig:case2}.

\begin{table*}[]

\begin{tabular}{p{1\linewidth}}
\hline
Given a query, this query may be ambiguous. For example, in this query, pronouns may be used to refer to entities or some components may be omitted, so you need to perform coreference resolution and ellipsis resolution. Please ask a question to clarify any unclear points in the query. You only need to output the clarification question, no need to output extra content. Here are some examples.\\
Examples:  \\
\textbf{\#Query\#}: Has she produced anything else? \\
\textbf{\#Clarification Question\#}: Who does "she" refer to? \\
\\
\textbf{\#Query\#}: Has she produced anything else? \\
\textbf{\#Clarification Question\#}: What does "anything else" exclude here? \\
\\
\textbf{\#Query\#}: Who were the first settlers? \\
\textbf{\#Clarification Question\#}: Where are the settlers referred to here? \\
\\
Please ask a clarification question about the following query. \\
\textbf{\#Query\#}: \{\texttt{Current Query}\} \\
\textbf{\#Clarification Question\#}:  \\
\hline
\end{tabular}
\caption{Prompt used in clarification question generation.}
\label{tab:clarification_prompt}
\end{table*}

\begin{table*}[]

\begin{tabular}{p{1\linewidth}}
\hline
Given a conversation and a clarification question, the final query in the conversation may be ambiguous. Please rephrase the final query based on the clarification question, address the issue raised, and do not change the original meaning. You only need to output the rephrased query without any extra content. Here are some examples.\\
Examples: \\
\textbf{\#Clarification Question\#}: \\
Who does "she" refer to?\\
\textbf{\#Conversation\#}: \\
Q: Who produced the original show one foot in the grave?\\
A: Susan Belbin.\\
Q: Has she produced anything else?\\
\textbf{\#Rewritten Query\#}: \\
Has susan belbin produced anything else?\\
\\
\textbf{\#Clarification Question\#}: \\
What does "anything else" exclude here?\\
\textbf{\#Conversation\#}: \\
Q: Who produced the original show one foot in the grave?\\
A: Susan Belbin.\\
Q: Has she produced anything else?\\
\textbf{\#Rewritten Query\#}: \\
Has she produced anything else besides one foot in the grave?\\
\\
\textbf{\#Clarification Question\#}: \\
Where are the settlers referred to here?\\
\textbf{\#Conversation\#}: \\
Q: Where was the indian ocean mentioned above located?\\
A: Indian Ocean is the third-largest of the world's oceanic divisions, it  is bounded by Asia to the north, Africa to the west and Australia to the east. To the south it is bounded by the Southern Ocean or Antarctica, depending on the definition in use. Along its core, the Indian Ocean has some large marginal or regional seas such as the Arabian Sea, the Laccadive Sea, the Somali Sea, Bay of Bengal, and the Andaman Sea.\\
Q: Who were the first settlers?\\
\textbf{\#Rewritten Query\#}: \\
Who were the first settlers of the indian ocean?\\
\\
Please rephrase the last query in the conversation based on the clarification question below.\\
\textbf{\#Clarification Question\#}: \\
\{\texttt{Clarification Question}\}\\
\textbf{\#Conversation\#}: \\
\{\texttt{Conversation}\}\\
\textbf{\#Rewritten Query\#}: \\
\hline
\end{tabular}
\caption{Prompt used in rewriting.}
\label{tab:rewriting_prompt}
\end{table*}

\end{document}